\documentclass{article}
\usepackage{ijcai24}

\usepackage{times}
\usepackage{soul}
\usepackage{url}
\usepackage[hidelinks]{hyperref}
\usepackage[utf8]{inputenc}
\usepackage[small]{caption}
\usepackage{graphicx}
\usepackage{amsmath}
\usepackage{amsthm}
\usepackage{booktabs}
\usepackage{algorithm}
\usepackage{algorithmic}
\usepackage[switch]{lineno}
\usepackage{bbm}
\newtheorem{problem}{Problem}
\newtheorem{remark}{Remark}
\usepackage{amsmath,amssymb,amsfonts}
\DeclareMathOperator*{\argmax}{arg\,max}

\usepackage{xcolor}

\title{A Behavior-Aware Approach for Deep Reinforcement Learning in Non-stationary Environments without Known Change Points}

\author{
Zihe Liu
\and
Jie Lu\and
Guangquan Zhang\And
Junyu Xuan
\affiliations
Australian Artificial Intelligence Institute (AAII), University of Technology Sydney
\emails
Zihe.Liu@student.uts.edu.au,
\{Jie.Lu, Guangquan.Zhang, Junyu.Xuan\}@uts.edu.au
}

\begin{document}

\maketitle

\begin{abstract}
Deep reinforcement learning is used in various domains, but usually under the assumption that the environment has stationary conditions like transitions and state distributions. When this assumption is not met, performance suffers. For this reason, tracking continuous environmental changes and adapting to unpredictable conditions is challenging yet crucial because it ensures that systems remain reliable and flexible in practical scenarios. Our research introduces Behavior-Aware Detection and Adaptation (BADA), an innovative framework that merges environmental change detection with behavior adaptation. The key inspiration behind our method is that policies exhibit different global behaviors in changing environments. Specifically, environmental changes are identified by analyzing variations between behaviors using Wasserstein distances without manually set thresholds. The model adapts to the new environment through behavior regularization based on the extent of changes. The results of a series of experiments demonstrate better performance relative to several current algorithms. This research also indicates significant potential for tackling this long-standing challenge.
\end{abstract}

\section{Introduction}

Deep reinforcement learning has extensive applications in economics~\cite{mosavi2020comprehensive}, energy engineering~\cite{delarue2020reinforcement,oikonomou2023hybrid}, medical analysis~\cite{hu2023reinforcement,tiwari2023consumer} and other domains, where policies are trained to make optimal sequential decisions in an assumed stationary environment. However, in practice, stationary environments are rare. Instead, the norm is non-stationary environments where the underlying environment can change in quite unpredictable and abrupt ways. For instance, outdoor robots must navigate constantly changing terrain and lighting levels, while financial markets should rapidly shift alongside breaking news and global events. Hence, ignoring the non-stationarity of underlying environments will frequently lead to poor performance even using a superior algorithm. There is no doubt that addressing this issue requires a dedicated strategy.
\begin{figure}[t]
\centering
\includegraphics[width=0.49\textwidth]{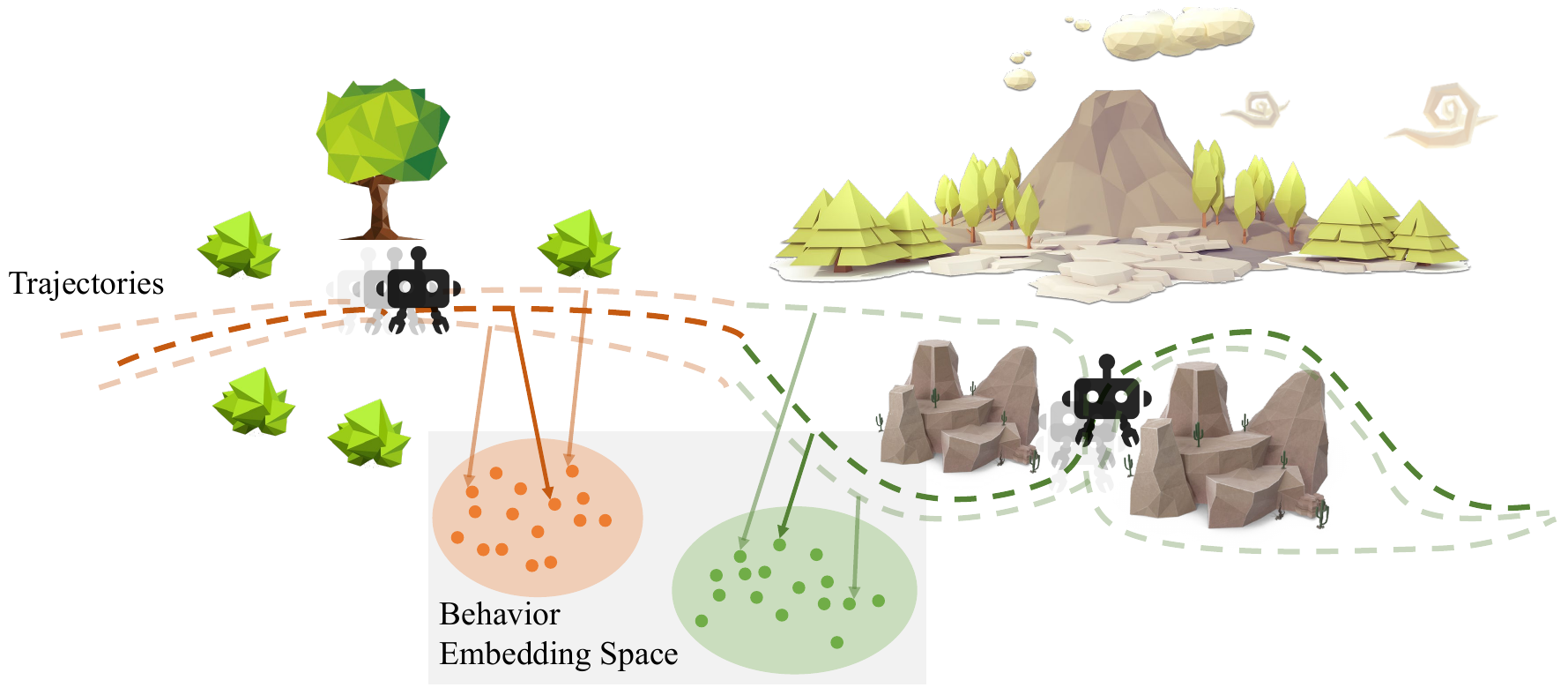}
\caption{When an outdoor robot moves from flat terrain to mountains, its speed, direction, and acceleration control changes corresponding to the changing conditions. We believe these variations can be fully captured through behavior.}
\label{fig:motivation}
\end{figure}

In prior work, several research teams have looked for solutions. Some have converted the problem into a continual multi-task reinforcement learning problem~\cite{kirkpatrick2017overcoming,schwarz2018progress}, while others have transformed the issue into a meta reinforcement learning problem~\cite{yu2020meta,xie2021deep}. Yet the common thread in all these studies is that the change points need to be known in advance, as these change points are used to divide the non-stationary environment into multiple tasks. However, there are often no ready-to-use indicators for unpredictable changes. Furthermore, a typical continual learning setting focuses on preventing catastrophic forgetting, while remembering the knowledge from previous tasks may not contribute to the current adaptation, especially in more practical environments without cyclically recurring tasks.
To address the absence of known change points, some research actively detects environmental changes using methods like reward-based detection~\cite {lomonaco2020continual,kirkpatrick2017overcoming} or state-based detection~\cite{padakandla2020reinforcement}. However, the reward-based method generally requires timely rewards and manually set thresholds. In addition, the state information is not comprehensive and accurate enough for detection because different global behaviors may have the same final state or perform similar actions at a local level~\cite{pacchiano2020learning}. Therefore, changes in state alone do not serve as reliable indicators for determining environmental changes.

We posit that the agents in an environment can be better characterized through their behavior. In our research, behavior represents the embeddings mapped from the sequences of states, actions and rewards during a period. As demonstrated in Fig.~\ref{fig:motivation}, when an outdoor robot encounters different environmental conditions, such as terrain, its speed and direction tend to demonstrate significant changes from those of the previous terrain. 
However, the separate variables like speed and direction at a few time steps can not describe the comprehensive trajectory change, making it challenging to understand and adapt to the new environment. In contrast, behavior can offer more comprehensive information from a global level. We believe that behavior distribution changes simultaneously reflect environmental changes and can help us adapt to new conditions, so our proposed method uses behavior as a core indicator and knowledge. 
We propose using these shifts in behavior distribution to detect environmental changes. Additionally, these changes indicate that departing from the behavior in the original environment is beneficial for optimal behaviors within the new conditions. 

Inspired by this, we present a novel approach to detect environment changes by monitoring behavior distribution shifts based on the Wasserstein distance~\cite{villani2009optimal,panaretos2019statistical}. The agent(s)’ behavior is then regularized accordingly to help the policy steer away from the previous optimum and adapt to new environmental conditions. 
Experiments in benchmark environments prove our method to be effective and accurate compared to other methods. 
We propose a setting that enhances the applicability and effectiveness of reinforcement learning across diverse fields, from robotics navigating in dynamic landscapes to trading systems that can respond to volatile markets. 

Our main contributions are summarized as follows,
\begin{itemize}
    \item We propose an environmental change detection method, testing environmental change points through the Wasserstein distance between the global behavior information without manually setting thresholds.
   \item With detected change points, we introduce a policy adaptation method that facilitates faster deviation from the previous optimum and exploration of new behavioral regions.
   We adjust regularization based on the extent of change to ensure adaptability under various conditions.
    \item We provide an end-to-end framework called Behavior-Aware Detection and Adaptation (BADA) to collaborate environment change detection and adaptation by analyzing and employing behavior.
\end{itemize}

\section{Related Work}

\paragraph{Change detection in RL.}~Several partial models~\cite{da2006dealing,hadoux2014sequential} have been published that represent environmental contexts using a quality signal, but neither method works well in complex scenarios.  
Online Parametric Dirichlet Change Point (ODCP)~\cite{prabuchandran2021change} detects environmental changes by converting data into unconstrained multivariate data. At the same time, CRL-Unsup\cite{lomonaco2020continual}  uses the gap between short and long-term rewards as an indicator, which relies on manually selected thresholds. Liu et al.~\cite{liu2024deep} detect the changes by analyzing the joint distribution of state and policy, lacking a comprehensive perspective over time.

\noindent\textbf{Adaptive/Transfer RL.}~Another feature of CRL-Unsup ~\cite{lomonaco2020continual} is that it adapts to new environments using elastic weight consolidation (EWC)~\cite{kirkpatrick2017overcoming}. In ODCP~\cite{padakandla2020reinforcement}, when a change point is detected, the Q value of the relevant model is used as an update parameter. Several approaches learn a latent representation incorporating shared and specific components from the source domain~\cite{huang2021adarl,zintgraf2019fast,trabucco2022anymorph}. Some work\cite{huang2021adarl,zintgraf2019fast,trabucco2022anymorph} learn a latent representation encompassing shared and specific components from source domains. These methods typically have clear task definitions and differentiate between the source and target domain. By contrast, our method, BADA, is designed for sequential changes and continuously adapts to new tasks.

\noindent\textbf{Continual RL.}~In continual scenarios, the focus tends to be placed on avoiding catastrophic forgetting, which is the tendency of a neural network to abruptly forget previously learned tasks upon learning a new task. Most schemes in continual learning are trained on pairs of separate tasks, and discrete transitions are often used to inform adaptations in the model. Some approaches add additional structures to the network model to resist forgetting~\cite{zenke2017continual,schwarz2018progress,aljundi2017expert} added additional structures to network models to resist forgetting. Other studies achieve this goal by introducing additional data or label inputs ~\cite{shin2017continual,GradientEpisodicMemory}. However, the change points of environments are usually unpredictable, which makes it difficult to deploy continual learning methods in such settings directly.

\noindent\textbf{Meta RL.} Meta reinforcement learning typically consists of meta-training and meta-testing, with the goal of learning a policy capable of adapting to new tasks from a given task distribution. Some methods learn latent variable models to infer the task embedding from current and past experiences. Off-policy reinforcement learning is then performed with this latent variable ~\cite{rakelly2019efficient,DBLP:journals/pami/BingLHK23,xie2021deep}. Unlike meta-reinforcement learning, where the training and testing tasks are separate, BADA focuses on real-time adaptation during the training phase. Moreover, BADA does not assume that the tasks come from one distribution.

\noindent\textbf{Multi-task RL.}
Multi-task learning within varied task families often faces the challenge of negative transfer among dissimilar tasks, which can impede training. Previous research addresses this issue by evaluating task relatedness using a validation loss for different tasks ~\cite{liu2022autolambda,fifty2021efficiently,standley2020tasks}.
Other complementary methods for sharing information include sharing data, parameters or representations, and sharing behaviors~\cite{yu2021conservative,yu2022leverage,sasaki2020behavioral,d2020sharing}. The goal with multi-task settings is to train an agent to perform well at various tasks simultaneously. By contrast, BADA focuses on adapting to the current environment for optimal performance.

\section{Methodology}
\subsection{Problem Formulation}
A Markov decision process $\mathbf{M}$ is defined by a state space $\mathcal{S}$, a starting state distribution $p_0(s)$, an action space $\mathcal{A}$, a transition dynamics $\mathcal{P}(s_{t+1}|s_t, a)$, and a reward function $\mathcal{R}: \mathcal{S} \times \mathcal{A} \rightarrow \mathbb{R}$. A policy $\pi_\theta$ is parameterized by $\theta$. The interaction trajectory $\tau = \{s_0, a_0, r_0, s_1, a_1, r_1,...\}$ is collected by a policy $\pi_\theta$. With 
a discount factor of $\gamma$, the optimal policy is the one that maximizes the expected discounted reward: $\pi^* = \argmax_{\pi} \mathbb{E}_{\tau \sim \pi}\left[ \sum_t \gamma^t \mathcal{R}(s_t, a_t) \right].$

Standard reinforcement learning assumes that the underlying $\mathbf{M}$ is unknown but fixed. When this assumption does not stand, a reinforcement learning scheme for non-stationary environments must be implemented. 
Further, this paper targets a specific problem within non-stationary environments,  in which the change happens suddenly, and the change points are unknown. Formally:

\begin{problem}

Let $\{\mathbf{M}_{k=1:K}\}$ be a sequence of different MDPs with unknown switch points $\{C_1, ..., C_{K-1}\}$ and arbitrary order. An agent will sequentially interact with $\{\mathbf{M}_{k=1:K}\}$ with unknown change points, where the goal is to find an optimal policy to maximize the long-term cumulative reward:
\begin{equation}
\begin{aligned}
	\pi^*_{1:J} = \argmax_{\pi_{1:J}}  \mathbb{E}_{\tau \sim {\pi}}\left[ \sum_{k=1:J} \sum_{t=C'_{k-1}}^{C'_k} \gamma^t \mathcal{R}(s_t, a_t) \right],
\end{aligned}
\end{equation} 
where $\{C'_1, ..., C'_{J-1}\}$ are the detected change points.
\end{problem}
\begin{remark}
Each MDP $\mathbf{M}_k$ is distinct, potentially differing in state spaces, transition dynamics, and reward functions.
\end{remark}
\begin{remark}
The duration of the agent's interaction in each MDP, $C_k - C_{k-1}$, is not predetermined and assumed.
\end{remark}
To ensure the maximum long-term reward, the problem encompasses two sub-goals: detecting change points accurately and adapting to the new environment rapidly.

\subsection{Behavior-based Change Detection}

During the training process, the policy continuously interacts with the environment. Within each update epoch $t$, the trajectories collected by $\pi_\theta$ are denoted as $\tau = \{s_0, a_0, r_0, ..., s_H, a_H, r_H\}$, where $H$ is the step taken in this epoch. 
A behavioral embedding map $\Phi: \Gamma \rightarrow \mathcal{E}$ maps the trajectories into a behavioral latent space. In our particular implementation, this map function is a multilayer perceptron. The embedding $\mathbb{P}_{\theta}$ represents the behavior embedding distribution corresponding to policy $\pi_\theta$ at epoch $t$.

As mentioned previously, environmental non-stationarity leads to a shift in the trajectory. Therefore, the behavior distributions from two adjacent epochs $\{\mathbb{P}_{t-1},\mathbb{P}_{\theta}\}$ are used to quickly identify the change points promptly. Here, the Wasserstein distance~\cite{olkin1982distance,panaretos2019statistical} is used as the measure for evaluating the difference between behavior trajectories. The Wasserstein distance originates from the optimal transport problem, which evaluates the cost required to transform one probability distribution into another.
Given two distributions $\mu, \nu$, the Wasserstein distance is defined as 
\begin{equation}
\mathrm{W}(\mu, \nu)=\inf _{\gamma \in \Gamma(\mu, \nu)} \int c(x, y) \mathrm{d} \gamma(x, y),
\label{wd}
\end{equation}
where $\Pi(\cdot, \cdot)$ denotes the joint distribution with marginal distributions, and $c(\cdot, \cdot)$ denotes the cost function quantifying the distance between two points. If the cost of a move is simply the distance between the two points, then the optimal cost is identical to the definition of the Wasserstein 1-distance~\cite{xu2019approximation}. We calculate the distance by using the dual form of Eq.~(\ref{wd}), which is defined as:
\begin{equation}
\mathrm{W}(\mu, \nu)=\sup _{f_\mu, f_\nu} \int f_\mu d \mu(x)-\int f_\nu d \nu(y) \quad, 
\label{dual-wd}
\end{equation}
where $f_\mu, f_\nu: \mathbb{R}^d \rightarrow \mathbb{R}$ and $\operatorname{Lip}(f_\mu) \leq 1$. The $\operatorname{Lip}(f)$ denotes the minimal Lipschitz constant for the function $f$. 
To calculate the Wasserstein distance, the objective is to find the optimal $f^*_\mu, f^*_\nu$ to maximize the integral.

\begin{figure}[t]
    \centering
    \includegraphics[width=0.49\textwidth]{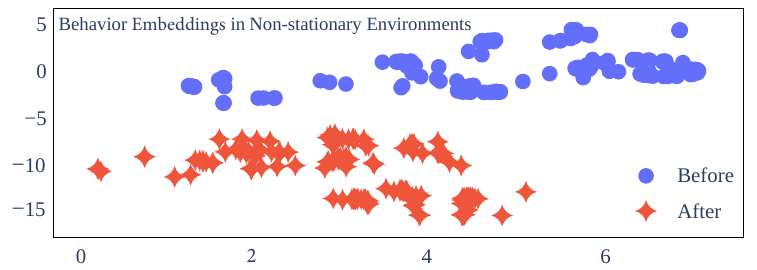}
    \caption{This figure presents a t-SNE plot of behavior. The distinct clusters demonstrate the significant impact of environmental changes on behavior and inspire us to use the behavior to adapt actively to coming changes.}
    \label{fig:tsne-traj}
\end{figure}

Wasserstein distance is a metric that reflects the proximity between two distributions, even if no overlap components exist. This property is important for our problem because the agent may manifest completely different behavior before and after changes. Therefore, the support between these distributions on behavior spaces would be limited, and then a proper distribution distance definition for such a situation is crucial. Additionally, its symmetrical nature offers a more effective measure of the differences between distributions compared with other options, like KL divergence.
For example, as Fig.~\ref{fig:tsne-traj} shows, when a policy is sequentially trained from one environment to another – say where the textures and lighting change – the agent’s behavior embedding distribution will show a distinct shift in distribution without overlapping. 
This observation can also help us identify these behavioral-level changes using the Wasserstein distance.

With the evaluated distance before and after a potential change point, we still need to decide on a change point, usually based on a manually determined threshold. It is difficult because it depends on the environment and behavior distributions, and what is even worse is that different change points may need different thresholds. Here, we propose to perform the permutation test \cite{welch1990construction,van2022comparing}, which infers the presence of any change points. The permutation test is an exact statistical hypothesis test based on proof by contradiction. This method involves permuting the order of samples, recalculating statistical test metrics, constructing an empirical distribution, and then determining the p-value based on this distribution to make inferences. 

To explain the permutation idea, given two samples from adjacent behavior embedding distributions $\mathbb{P}_{\theta}, \mathbb{P}_{t-1}$ and calculate the test statistic $T = \mathrm{W}(\mathbb{P}_{\theta}, \mathbb{P}_{t-1})$. The typical null hypothesis is given by:
\begin{equation}
    H_0: \mathbb{P}_{\theta}=\mathbb{P}_{t-1},
\end{equation}
i.e., all samples come from the same distribution. Then, for each permutation $e = 1, 2, ...E$, randomly permute the components of $\mathbb{P}_{\theta} \cup \mathbb{P}_{t-1}$, and split the permuted data into $\mathbb{P}_{\theta}^{(e)}, \mathbb{P}_{t-1}^{(e)}$ with the original sizes, then calculate test statistics $T_e=\mathrm{W}(\mathbb{P}_{\theta}^{(e)}, \mathbb{P}_{t-1}^{(e)})$. By repeating the permutation and calculation, a p-value is given by
\begin{equation}
    p=\frac{1}{E}\sum^{E}_{t=1}1\{T_e\geq T\},
\end{equation}
where $1$ is an indicator function. This test is guaranteed to control the type-I error~\cite{good2013permutation} because we evaluate the p-value of the test via the permutation approach. In addition, the non-parametric nature, i.e., that it does not rely on assumptions about data distribution. As Fig.~\ref{fig:tsne-traj} shows, the trajectory distribution usually does not conform to an easily computable and representable form of distribution. Therefore, using a permutation test is highly suitable for solving our problem. Suppose the p-value is lower than the significance level; in that case, the current epoch $t$ is noted as a change point $c=t$, and $\mathbb{P}_{pre}=\mathbb{P}_{c-1}$ is the optimal behavior distribution corresponding to the previous environment.

\begin{figure}[t]
    \centering
    \includegraphics{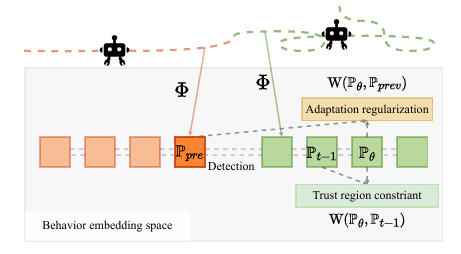}
    \caption{The BADA framework. When a change is detected through the behavior distribution permutation test, regularization will be added to deviate policy behavior from the previous optimum.}
    \label{fig:framework}
\end{figure}

\subsection{Behavior-Aware Adaptation}
Although the vanilla DRL can adapt to the new environment, especially in gradually changing environments, it typically requires many interactions that sample inefficient and generate a significant delay. With the detection signal from the above section, we aim to achieve fast adaption. Since our detection is based on Wasserstein distance, we follow Wasserstein-based policy gradient baseline - Behavior Guided Policy Gradients (BGPG)~\cite{pacchiano2020learning}. Its training objective (for stationary environment) is to maximize:
\begin{equation}
F(\theta) = \mathbb{E}_{\tau \sim \mathbb{P}_{\theta}}\left[\mathcal{R}(\tau)\right] - W(\mathbb{P}_{\theta},\mathbb{P}_{t-1}),
\label{eq:basic}
\end{equation}
where $\mathbb{P}_{t-1}$ is the behavior distribution of last update epoch.
\begin{algorithm}[tb]
    \caption{Behavioral Aware Detection and Adaptation (BADA)}
    \label{algo:all}
    \textbf{Initialize}: Policy $\pi_\theta$, behavioral embedding mapping function $\Phi$, and significance level $\alpha$.
    
    \begin{algorithmic}[1] 
        \FOR{Epoch $t = 1,2,...$}
        \STATE Collect $\tau = \{s_0, a_0, r_0, ..., s_H, a_H, r_H\}$ from the current environment.
        \STATE Obtain behavior embedding $\mathbb{P}_{\theta}$ by behavioral embedding mapping function $\Phi$.
        \STATE Compute the original statistics $T=W(\mathbb{P}_{\theta},\mathbb{P}_{t-1})$.
        \FOR{Permute iteration $e = 1,2,...E$}
        \STATE Shuffle $\mathbb{P}_{\theta} \cup \mathbb{P}_{t-1}$ and split the data into $\mathbb{P}_{\theta}^{(e)}, \mathbb{P}_{t-1}^{(e)}$ and compute statistics $T_e=W(\mathbb{P}_{\theta}^{(e)},\mathbb{P}_{t-1}^{(e)})$.
        \ENDFOR
        \STATE Obtain the p-value $\frac{1}{E}\sum^{E}_{t=1}1\{T_e\geq T\}$
        \IF {p-value$\leq \alpha$ at epoch $c$}
        \STATE Save $\mathbb{P}_{c-1}$ as previous behavior distribution $\mathbb{P}_{pre}$.
        \STATE Update policy parameter by $\theta_{t+1} \leftarrow  \theta_t - \alpha \nabla_\theta F(\theta_t)$  following Eq.~(\ref{eq:our})
        \ELSE
        \STATE Update policy parameter by $\theta_{t+1} \leftarrow  \theta_t - \alpha \nabla_\theta F(\theta_t)$ following Eq.~(\ref{eq:basic}).
        \ENDIF
        \STATE Save $\mathbb{P}_{t-1} \leftarrow \mathbb{P}_{\theta}$ for environment change detection.
        \ENDFOR
    \end{algorithmic}
\end{algorithm}

When a policy converges in one environment, the behavior will enter a relatively stable distribution, providing a basis for us to detect environmental changes. When a change point is detected at epoch $c$, indicating that a significant change in the environment has occurred, $\mathbb{P}_{c-1}$ is saved as a previous optimal behavior distribution $\mathbb{P}_{pre}$. 
To assist the policy in quickly deviating from the optimal behavior of the previous environment, we propose to add a regularizer that maximizes the difference between the current behavior distribution and the previously converged behavior distribution. This new objective function is designed as follows:
\begin{equation}
F(\theta) = \mathbb{E}_{\tau \sim \mathbb{P}_{\theta}}\left[\mathcal{R}(\tau)\right] - \mathrm{W}(\mathbb{P}_{\theta},\mathbb{P}_{t-1}) + \delta \mathrm{W}(\mathbb{P}_{\theta},\mathbb{P}_{pre}),
\label{eq:our}
\end{equation}
where $\mathcal{R}=\sum A^{\pi_{t-1}}\left(s_i, a_i\right) \frac{\pi_\theta\left(a_i \mid s_i\right)}{\pi_{t-1}\left(a_i \mid s_i\right)}$, $ A^{\pi_{t-1}}\left(s_i, a_i\right) $ is the advantage function, and $\mathbb{P}_{pre}$ is the converged behavior distribution in the previous environment, and $\delta  \in \mathbb{R}_{>0}$ is a hyper-parameter. 
Here, we use the adjacent behavior distance on the detected change point $W(\mathbb{P}_{c-1},\mathbb{P}_{c})$ as $\delta$, depending on the extent of change. This self-adjusted coefficient ensures that the adaptation regularization has a greater impact as the level of environmental change increases.

If no change is detected, the adaptation term will not work, so $\delta$ will be set as zero. The first penalty constrains policy updates within a trust region, ensuring the validity of importance sampling. However, this constraint can lead to slow adaptation when the environment undergoes abrupt changes, as the policy hesitates to deviate from its previous optimal behavior. At the change point $c$, $\mathbb{P}_{prev} = \mathbb{P}_{c-1}$. Only following the first penalty term at this point might trap the policy in a suboptimal area for an extended period. Therefore, our second adaptation regularization serves as a contrastive term, steering the policy away from previous behavior. As the policy gradually adapts to the current environment, i.e., $t \gg c$, the adaptation term $\mathrm{W}(\mathbb{P}_{\theta},\mathbb{P}_{c-1})$ and the first term $\mathrm{W}(\mathbb{P}_{\theta},\mathbb{P}_{t-1})$ no longer conflict. The penalty constraints ensure performance improvement in a stationary environment, and the role of the adaptation term weakens as the policy moves away from the previous optimum.

With the optimal $f^*_\mu, f^*_\nu$ according to Eq.~(\ref{dual-wd}), the regularization term in Eq.~(\ref{eq:our}) is:
\begin{equation}
\mathrm{W}(\mathbb{P}_{\theta},\mathbb{P}_{pre}) \approx \mathbb{E}_{\tau \sim \mathbb{P}_{\theta}}[f^*_\mu(\tau)] -\mathbb{E}_{\phi \sim \mathbb{P}_{pre}}[f^*_\mu(\phi)].
\end{equation}
Maximizing this term can guide the optimization by favoring those trajectories that show more difference between old ones. When another change occurs, we consider only the preceding behavior distribution. We believe excessive constraints may lead to a narrow area and result in local optima. Therefore, focusing on the immediate historical behavior ensures adaptability to changing environments without introducing unnecessary complexities. This training goal allows us to scale to scenarios with multiple changes easily.
Fig.~\ref{fig:framework} illustrates the adaptation scheme, and Algorithm~\ref{algo:all} describes the complete BADA method in detail.

\section{Experiments and Analysis}
This section comprehensively evaluates our BADA method, addressing key questions: 1) Can BADA achieve higher rewards in environments without known change points? 2) Is behavior-based change detection superior to alternative methods? 3) Does BADA's adaptation method outperform retraining and other adaptation approaches? 4) Can BADA maintain performance with frequent environmental changes? These inquiries guide our experiments and analysis.

\subsection{Settings}
\begin{figure}[t]
    \centering
    \includegraphics[width=\columnwidth]{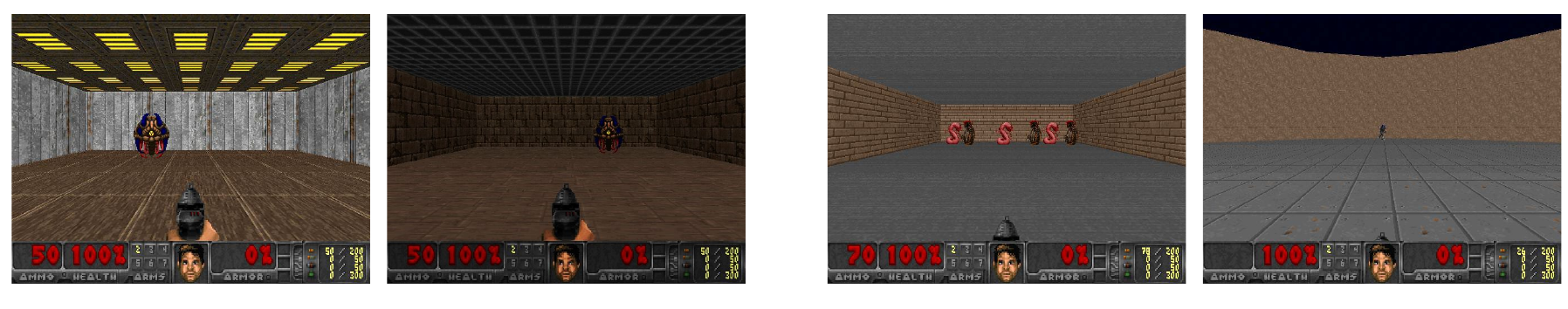}
    \caption{The simulated non-stationary environments. The left setting is from high-contrast \textit{simpler\_basic} to dimly lit \textit{basic} scenario, and the right one is from \textit{defend\_the\_line} with a rectangular map to \textit{defend\_the\_center} with a circular map.}
    \label{fig:env-setting}
\end{figure}

\textbf{Environments.}
We conducted all experiments within ViZDoom~\cite{Wydmuch2019ViZdoom}, a first-person shooting game with various scenarios. This environment allows reinforcement learning agents to be developed using only visual information (the screen buffer). We chose four scenarios to evaluate our proposed method. We employ distinct challenges and modifications to simulate dynamic environments for agent training. For example, as Fig.~\ref{fig:env-setting} shows,  the environment transit from high-contrast \textit{simpler\_basic} to dimly lit \textit{basic} settings, shift from defending a line in a rectangle map to defending a point in a circular map against enemies in \textit{defend\_the\_line/center}. In addition, we adjust the number of enemies in \textit{deadly\_corridor} and change the medikit textures in the \textit{healt\_gathering} scenario to represent new rooms. Agents need to respond to these changes.

\noindent\textbf{Comparison methods.}
In all experiments, we used the PPO/TRPO update, and once the environment changed, the model could not access any information on the changed environmental conditions. Further, we compared BADA to three baseline methods as follows.
\begin{itemize}
     \item PPO~\cite{schulman2017proximal} and TRPO~\cite{schulman2015trust} without detection and adaptation;
     \item Behavior-based BGPG~\cite{pacchiano2020learning} without detection and adaptation;
     \item CRL-Unsup~\cite{lomonaco2020continual} with both detection and adaptation.
 \end{itemize}
The agent architecture for all methods consisted of a 4-layer convolutional neural network (ConvNet) with 3x3 kernels featuring 16 maps, complemented by ReLU activation functions. This was followed by a fully connected layer that outputs a distribution of action sizes.

To evaluate performance in terms of environmental change detection, we compared BADA to:
\begin{itemize}
    \item A permutation test using KL divergence
    \item A two-sample test using weighted maximum mean discrepancy (WMMD) \cite{pmlr-v161-bellot21b}
    \item The online parametric Dirichlet change point (ODCP) \cite{prabuchandran2021change}
    \item  CRL-Unsup~\cite{lomonaco2020continual}, which is based on long and short-term rewards.
\end{itemize}

\begin{figure*}[t]
     \centering
      \includegraphics[width=\linewidth]{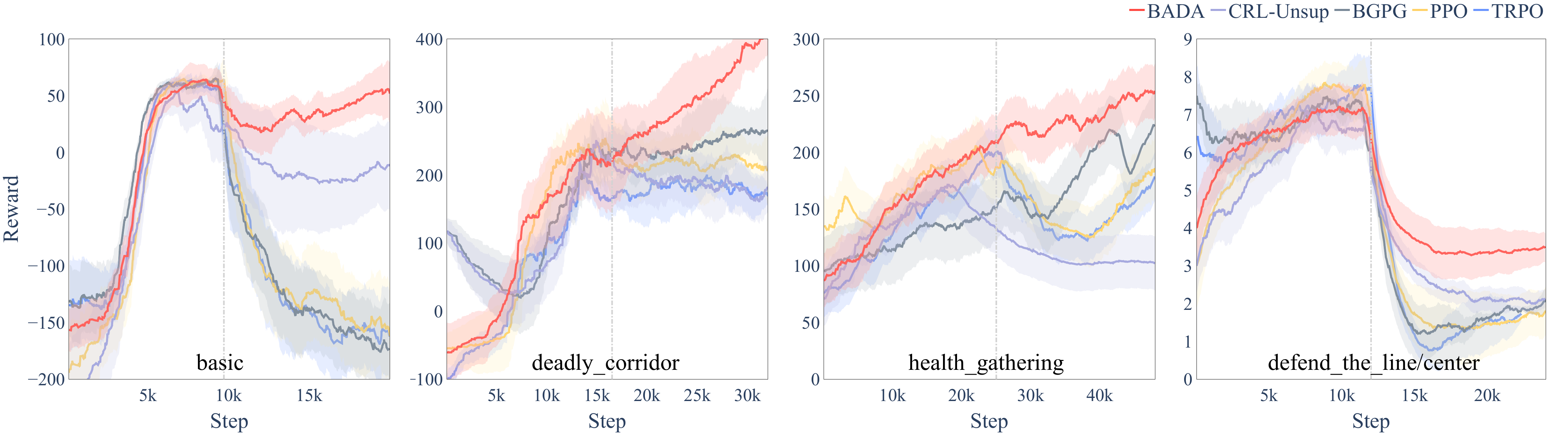}
        \caption{Performance comparison of different methods in non-stationary environments. The vertical dashed lines represent the points of environmental change, and the shaded areas around the reward lines indicate the standard deviation over different runs.}
        \label{fig:overall}
\end{figure*}

\textbf{Metrics.}
One metric is the cumulative reward or reward curve, and the other metric is F1 Score $\mathrm{F}_1=\frac{2 * P * R}{P+R}$~\cite{sasaki2007truth}, indicating the detection accuracy.

\subsection{Overall performance}

\noindent\textbf{Cumulative rewards.}~As Fig.~\ref{fig:overall} shows, BADA (depicted in red) exhibits an accelerated increase in reward after the change point (marked by the vertical dashed line in each graph). The post-change point improvement in reward is not only attributed to the effectiveness of adaptation regularization, enabling the policy to deviate from its previous optimum swiftly but also indicates that BADA accurately responds to environmental changes.

In \textit{basic}, when the lighting and wall texture of the room change, methods without adaptation see a significant performance drop. We can see that the CRL-Unsup method demonstrates a notable adaptation ability with a steady increase in rewards after environmental changes are initially detected, albeit slightly inferior to BADA. This indicates that our behavior-based regularization term enables faster adaptation to new environments.
In \textit{health\_gathering}, we can see that the texture of the medkit has a lesser impact than the lighting level. This is evident from the results, where even methods without adaptation can regain relatively high rewards after several updates. Meanwhile, BADA's reward continues to rise, demonstrating its adaptation capability in environments with relatively minor changes.
In this scenario, CRL-Unsup seems to be prone to false detections of environmental changes. This leads to unnecessary adaptations and results in a less stable learning process. 
Notably, in the \textit{deadly\_corridor}, the reduction in the number of enemies does not yield additional rewards for well-trained agents who do not employ adaptation strategies. This could be attributed to the fact that they still follow their original behavior and fail to respond promptly to environmental changes. However, BADA reaches a higher reward faster and sustains an upward trend, showing its ability to learn from new environmental conditions continuously. 
In the scenarios of \textit{defend\_the\_line/center}, map shape and defended goal changes force each method to learn a new task. In this context, BADA and CRL-Unsup outperform other baselines by quickly achieving higher scores on the new task, while BADA has superior performance compared to other methods. This indicates the effectiveness of steering away from the previous optimal strategy in discovering a new one.

\begin{table}[t]
    \centering

    \begin{tabular}{lll}
        \toprule
          &\textit{basic} & \textit{health\_gathering} \\
        \midrule
        BADA(Ours)     &\textbf{0.95$\mathbf{\pm0.08}$} &\textbf{0.90$\mathbf{\pm0.11}$}  \\
        Permu-KL    & 0.70$\pm0.09$ &0.50$\pm0.26$      \\
        CRL-Unsup    & 0.80$\pm0.12$ &0.35$\pm0.16$      \\
        WMMD & 0.50$\pm0.13$          &     0.56$\pm0.27$   \\
        ODCP & 0.55$\pm0.36$          & 0.37$\pm0.07$       \\
                \toprule
    &\textit{deadly\_corridor}&\textit{defend\_the\_line}\\
        \midrule
        BADA(Ours)       & \textbf{0.78$\mathbf{\pm0.16}$} & \textbf{0.86$\mathbf{\pm0.07}$}    \\
        Permu-KL    & 0.69$\pm0.09$ &0.50$\pm0.26$      \\
        CRL-Unsup    & 0.67$\pm0.21$ &0.72$\pm0.11$      \\
        WMMD & 0.47$\pm0.19$&0.62$\pm0.15$   \\
        ODCP& 0.38$\pm0.20$&0.50$\pm0.19$        \\
        \bottomrule
    \end{tabular}
    \caption{Comparative F1 scores of change detection methods in non-stationary environments.}
    \label{tab:detection-all}
\end{table}

\noindent\textbf{Environment change detection accuracy.}
Tab.~\ref{tab:detection-all} lists the F1 scores for all the methods. As shown, BADA outperforms other methods in all scenarios. 
The lower accuracy of BADA in the \textit{deadly\_corridor} scenario is because of the reduced number of enemies, which has a less immediate impact on behavior compared to observable conditions like environmental lighting.
The relatively poorer performance of the permutation test based on KL divergence (Permu-KL) compared to our Wasserstein-based approach can be attributed to the KL divergence being sensitive to the probability distribution’s representation in the data. In scenarios where the probability distributions of the environment states are sparse or have non-overlapping supports, KL divergence struggles to measure the distance between distributions accurately. By contrast, Wasserstein distance is based on the optimal transport problem, denoting the minimum "cost" of turning one distribution into the other. It is particularly beneficial in non-stationary reinforcement learning environments, which often feature abrupt and significant changes in distribution.
Another observation is that CRL-Unsup performs relatively well but heavily relies on extensive tests to select the thresholds manually. By contrast, BADA detection does not require hyperparameters to be manually adjusted and provides a significance level at the same time. We also find that ODCP and WMMD are not efficient in image-based scenarios.

\begin{figure}[t]
     \centering
        \includegraphics[width=\columnwidth]{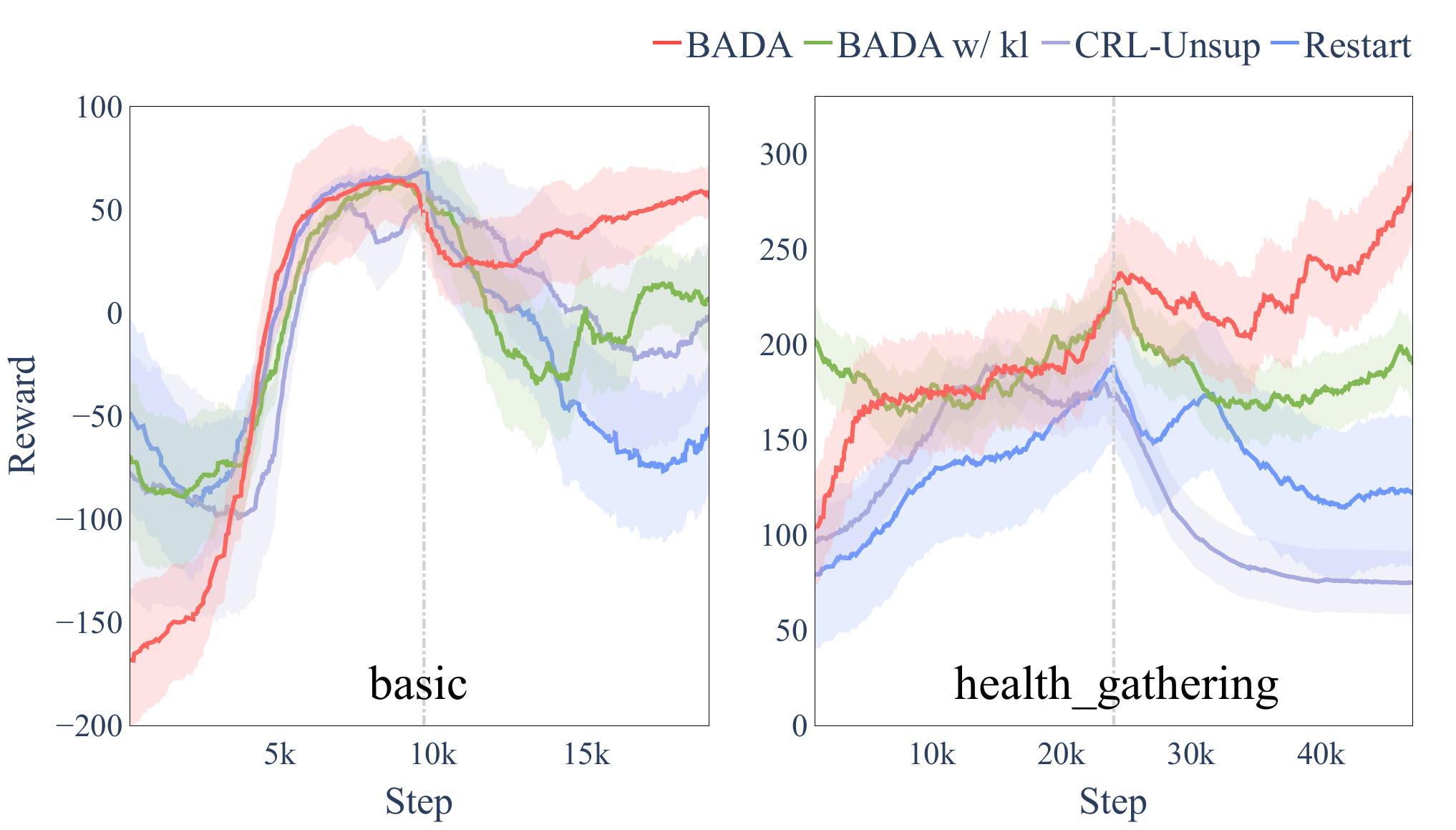}
   
        \caption{Cumulative rewards of adaptation strategies in non-stationary environments with known change points.}
        \label{fig:adapataion}
\end{figure}

\subsection{Ablation Study}\label{ablation-study}
\noindent\textbf{Adaptation evaluation.}
To evaluate the performance of adaptation separately and confirm whether the adaptation scheme contributes to new training as opposed to simply retraining the agent following a reinforcement learning loss, we test the following comparison methods:
\begin{itemize}
    \item Employing KL divergence instead of Wasserstein distance as the regularization term.
    \item CRL-Unsup with the EWC adaptation method.
    \item Restarting training following a traditional PPO scheme.
\end{itemize}
All methods are provided with the change points to initiate adaptation or retraining.

As shown in Fig.~\ref{fig:adapataion}, the BADA method excels in adapting to environments with known change points. First, BADA consistently outperformed the ‘Restart’ approach, as seen by the quicker recovery and sustained improvement in rewards. This indicates that BADA deviates from the previous optimal and finds a new one rapidly, adapting a more efficient strategy than restarting training from scratch. Second, BADA surpassed the other adaptation strategies, with the Wasserstein Distance constraint proving superior to KL divergence. Further, BADA outperforms CRL-Unsup, demonstrating that our behavior-based adaptation is more effective than other methods. Overall, these results confirm that BADA has a superior ability to adapt to environmental changes.

\begin{figure}[t]
    \centering
    \includegraphics[width=0.49\textwidth]{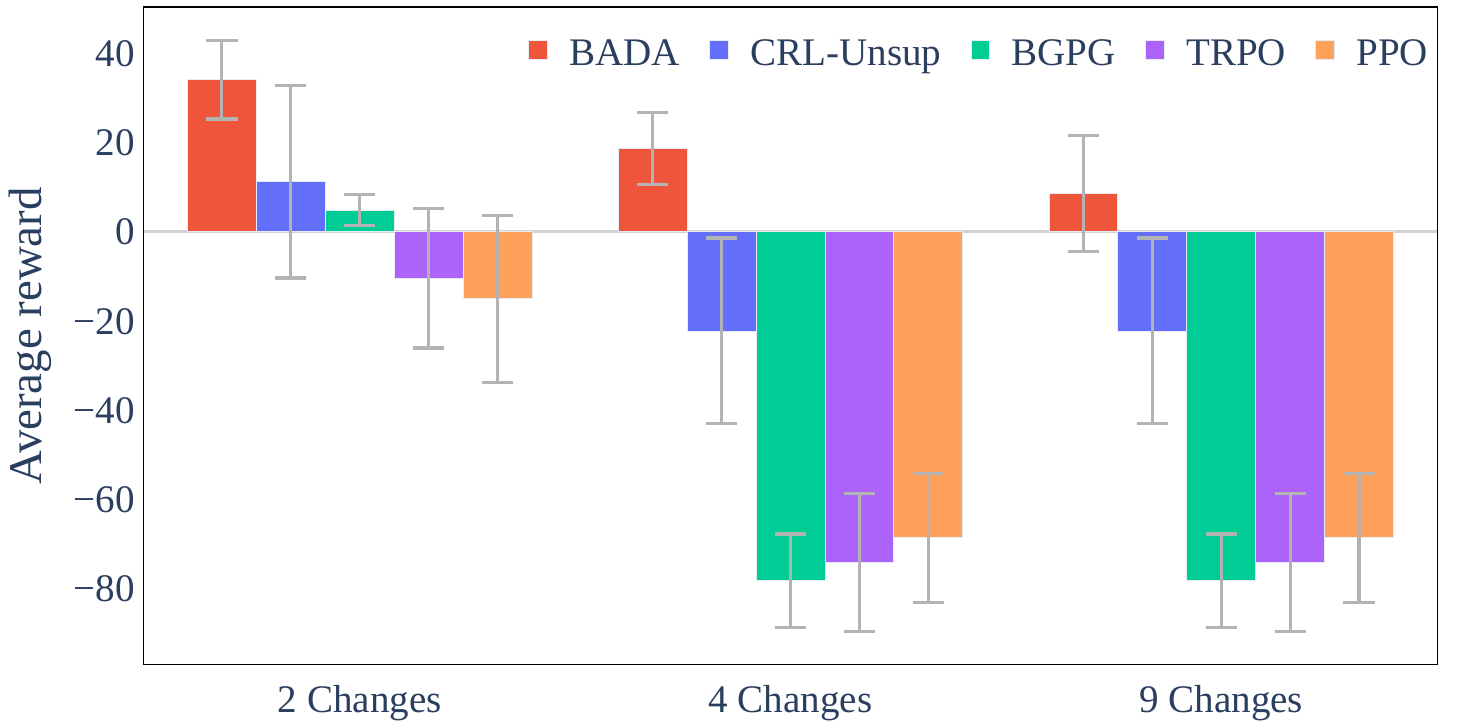}
    \caption{Average reward after the first change points in environments with increasing change points.}
    \label{fig:freq-change}
\end{figure}

\noindent\textbf{Frequently changing environments.}
Change frequency could challenge BADA’s ability because it may affect whether the constrained distribution is the previous optimal, i.e., the policy might not have converged when the changes occurred. 
Fig.~\ref{fig:freq-change} provides a comparative overview of the different algorithms’ performances across the \textit{basic} environment with varying numbers of change points. As indicated in red, BADA consistently achieved higher average rewards than the other methods when the environment changed in more frequently changing environments, demonstrating its robustness in dealing with multiple change points.
We can see that when the number of change points increases from 2 to 4, performance does not drop significantly. However, as the number of change points increases to 9, the performance of all methods has marked declines. However, BADA shows the most minor decrease, maintaining a clear lead over the others. This indicates BADA’s superior adaptability in more complex environments with frequent changes. Also, Tab.~\ref{tab:detection-freq} shows the detection accuracy in frequently changing environments. All methods will be influenced as the number of change points increases. Therefore, in extremely non-stationary environments, the policy may not have converged in each environment, resulting in behaviors that remain in random and chaotic patterns. This can limit BADA's ability to detect and adapt based on behavior.

\begin{table}[t]
    \centering

    \begin{tabular}{llll}
        \toprule
          &2 changes& 4 changes &9 changes\\
        \midrule
        BADA(Ours)     &\textbf{0.89$\mathbf{\pm0.12}$} &\textbf{0.78$\mathbf{\pm0.16}$}   & \textbf{0.56$\mathbf{\pm0.20}$} \\
        Permu-KL  & 0.52$\pm0.19$          & 0.49$\pm0.11$   & 0.43$\pm0.09$    \\
        CRL-Unsup  & 0.71$\pm0.13$          & 0.60$\pm0.17$   & 0.37$\pm0.08$    \\
        WMMD & 0.45$\pm0.17$          &     0.40$\pm0.11$ & 0.28$\pm0.19$\\
        ODCP & 0.25$\pm0.15$          & 0.21$\pm0.11$ & 0.20$\pm0.13$\\
        \bottomrule
    \end{tabular}
        \caption{F1 scores for change detection methods across environments with increasing change points.}
    \label{tab:detection-freq}
\end{table}

\begin{figure}[t]
    \centering
\includegraphics[width=\columnwidth]{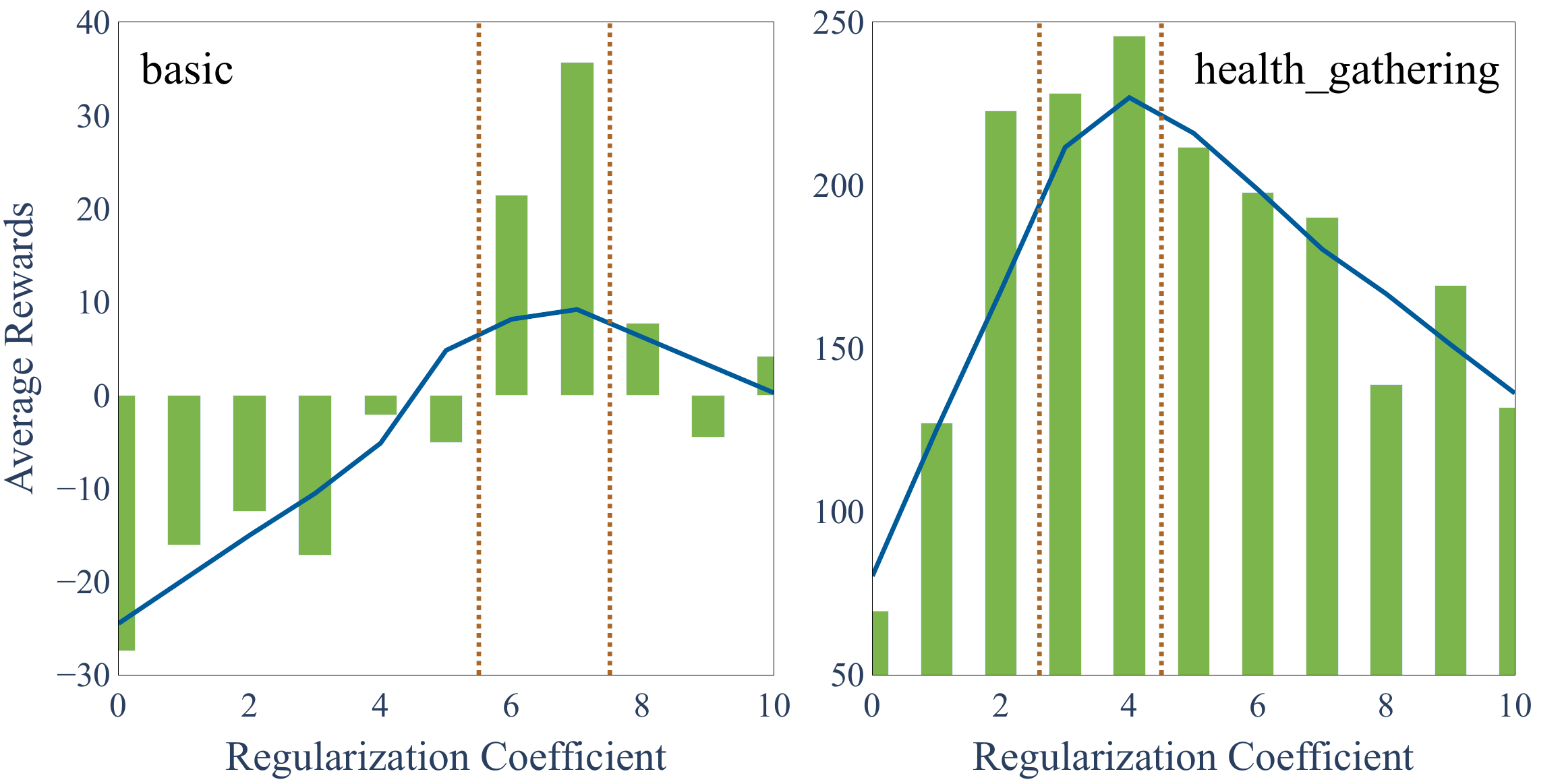}

    \caption{The parameter sensitivity analysis of the adaptation regularization. The orange lines represent the coefficient range we used.}
    \label{fig:para}
\end{figure}

\noindent\textbf{Parameter sensitivity.}
Fig.~\ref{fig:para} shows the parameter sensitivity analysis for the adaptation regularization term $\mathrm{W}(\mathbb{P}_{\theta},\mathbb{P}_{pre})$ in Eq.~(\ref{eq:our}). The result indicates a distinct peak in average rewards for both environments. This peak represents an optimal value for the coefficient, consistent with the range of our adaptive coefficient (denoted in orange rectangles). The parameter we used represents the extent of environmental change, determining the level of adaptation based on the environment. The experiment proves that our self-adjusted coefficient, according to environmental changes, is effective. Also, the decline post-peak implies that an overly aggressive correction term could negatively impact the learning process. Therefore, an accurate balance of adaptation regularization is crucial to sustaining good performance. The empirical results show that tuning it according to change level is valid.

\section{Conclusion}\label{conclusion}

This paper addresses deep reinforcement learning in non-stationary environments without known change points by developing the Behavior-Aware Detection and Adaptation (BADA) framework. The behavior-based change detection method represents a novel approach to monitoring and responding to environmental shifts by closely analyzing policy behavior. This method has proven effective and accurate without any manually set threshold, allowing for timely adjustments to the learning strategy. Furthermore, the online adaptation mechanism integrates this behavioral information, providing a self-adjusted regularization term. The behavior-based regularization can help policy steer from suboptimal areas and find potential behavior in new conditions. The experimental results show its superior performance in accurately detecting changes and quickly adapting to new environments compared to other methods. A future extension could benefit from exploring mechanisms for off-policy adaptations, broadening BADA's applicability in various RL settings.

\clearpage
\section*{Acknowledgements}
This work is supported by the Australian Research Council under Australian Laureate Fellowships FL190100149 and Discovery Early Career Researcher Award DE200100245.

\bibliographystyle{named}
\bibliography{ijcai24}

\begin{thebibliography}{}

\bibitem[\protect\citeauthoryear{Aljundi \bgroup \em et al.\egroup }{2017}]{aljundi2017expert}
Rahaf Aljundi, Punarjay Chakravarty, and Tinne Tuytelaars.
\newblock Expert gate: Lifelong learning with a network of experts.
\newblock In {\em IEEE Conference on Computer Vision and Pattern Recognition (CVPR)}, pages 3366--3375, 2017.

\bibitem[\protect\citeauthoryear{Bellot and van~der Schaar}{2021}]{pmlr-v161-bellot21b}
Alexis Bellot and Mihaela van~der Schaar.
\newblock A kernel two-sample test with selection bias.
\newblock In Cassio de~Campos and Marloes~H. Maathuis, editors, {\em Proceedings of the Thirty-Seventh Conference on Uncertainty in Artificial Intelligence}, volume 161 of {\em Proceedings of Machine Learning Research}, pages 205--214, 2021.

\bibitem[\protect\citeauthoryear{Bing \bgroup \em et al.\egroup }{2023}]{DBLP:journals/pami/BingLHK23}
Zhenshan Bing, David Lerch, Kai Huang, and Alois~C. Knoll.
\newblock Meta-reinforcement learning in non-stationary and dynamic environments.
\newblock {\em {IEEE} Trans. Pattern Anal. Mach. Intell.}, 45(3):3476--3491, 2023.

\bibitem[\protect\citeauthoryear{Da~Silva \bgroup \em et al.\egroup }{2006}]{da2006dealing}
Bruno~C Da~Silva, Eduardo~W Basso, Ana~LC Bazzan, and Paulo~M Engel.
\newblock Dealing with non-stationary environments using context detection.
\newblock In {\em International conference on Machine learning (ICML)}, pages 217--224, 2006.

\bibitem[\protect\citeauthoryear{Delarue \bgroup \em et al.\egroup }{2020}]{delarue2020reinforcement}
Arthur Delarue, Ross Anderson, and Christian Tjandraatmadja.
\newblock Reinforcement learning with combinatorial actions: An application to vehicle routing.
\newblock {\em Advances in Neural Information Processing Systems}, 33:609--620, 2020.

\bibitem[\protect\citeauthoryear{D'Eramo \bgroup \em et al.\egroup }{2020}]{d2020sharing}
Carlo D'Eramo, Davide Tateo, Andrea Bonarini, Marcello Restelli, Jan Peters, et~al.
\newblock Sharing knowledge in multi-task deep reinforcement learning.
\newblock In {\em 8th International Conference on Learning Representations,$\{$ICLR$\}$ 2020, Addis Ababa, Ethiopia, April 26-30, 2020}, pages 1--11. OpenReview. net, 2020.

\bibitem[\protect\citeauthoryear{Fifty \bgroup \em et al.\egroup }{2021}]{fifty2021efficiently}
Chris Fifty, Ehsan Amid, Zhe Zhao, Tianhe Yu, Rohan Anil, and Chelsea Finn.
\newblock Efficiently identifying task groupings for multi-task learning.
\newblock {\em Advances in Neural Information Processing Systems}, 34:27503--27516, 2021.

\bibitem[\protect\citeauthoryear{Good}{2013}]{good2013permutation}
Phillip Good.
\newblock {\em Permutation tests: a practical guide to resampling methods for testing hypotheses}.
\newblock Springer Science \& Business Media, 2013.

\bibitem[\protect\citeauthoryear{Hadoux \bgroup \em et al.\egroup }{2014}]{hadoux2014sequential}
Emmanuel Hadoux, Aur{\'e}lie Beynier, and Paul Weng.
\newblock Sequential decision-making under non-stationary environments via sequential change-point detection.
\newblock In {\em Learning over Multiple Contexts (LMCE)}, 2014.

\bibitem[\protect\citeauthoryear{Hu \bgroup \em et al.\egroup }{2023}]{hu2023reinforcement}
Mingzhe Hu, Jiahan Zhang, Luke Matkovic, Tian Liu, and Xiaofeng Yang.
\newblock Reinforcement learning in medical image analysis: Concepts, applications, challenges, and future directions.
\newblock {\em Journal of Applied Clinical Medical Physics}, 24(2):e13898, 2023.

\bibitem[\protect\citeauthoryear{Huang \bgroup \em et al.\egroup }{2022}]{huang2021adarl}
Biwei Huang, Fan Feng, Chaochao Lu, Sara Magliacane, and Kun Zhang.
\newblock Adarl: What, where, and how to adapt in transfer reinforcement learning.
\newblock In {\em International Conference on Learning Representations (ICLR)}, 2022.

\bibitem[\protect\citeauthoryear{Kirkpatrick \bgroup \em et al.\egroup }{2017}]{kirkpatrick2017overcoming}
James Kirkpatrick, Razvan Pascanu, Neil Rabinowitz, Joel Veness, Guillaume Desjardins, Andrei~A Rusu, Kieran Milan, John Quan, Tiago Ramalho, Agnieszka Grabska-Barwinska, et~al.
\newblock Overcoming catastrophic forgetting in neural networks.
\newblock {\em Proceedings of the National Academy of Sciences}, 114(13):3521--3526, 2017.

\bibitem[\protect\citeauthoryear{Liu \bgroup \em et al.\egroup }{2022}]{liu2022autolambda}
Shikun Liu, Stephen James, Andrew Davison, and Edward Johns.
\newblock Auto-lambda: Disentangling dynamic task relationships.
\newblock {\em Transactions on Machine Learning Research}, 2022.

\bibitem[\protect\citeauthoryear{Liu \bgroup \em et al.\egroup }{2024}]{liu2024deep}
Zihe Liu, Jie Lu, Junyu Xuan, and Guangquan Zhang.
\newblock Deep reinforcement learning in nonstationary environments with unknown change points.
\newblock {\em IEEE Transactions on Cybernetics}, 2024.

\bibitem[\protect\citeauthoryear{Lomonaco \bgroup \em et al.\egroup }{2020}]{lomonaco2020continual}
Vincenzo Lomonaco, Karan Desai, Eugenio Culurciello, and Davide Maltoni.
\newblock Continual reinforcement learning in 3d non-stationary environments.
\newblock In {\em IEEE/CVF Conference on Computer Vision and Pattern Recognition Workshops (CVPRW)}, pages 248--249, 2020.

\bibitem[\protect\citeauthoryear{Lopez-Paz and Ranzato}{2017}]{GradientEpisodicMemory}
David Lopez-Paz and Marc'Aurelio Ranzato.
\newblock Gradient episodic memory for continual learning.
\newblock In {\em Advances in Neural Information Processing Systems (NIPS)}, pages 6467--6476, 2017.

\bibitem[\protect\citeauthoryear{Mosavi \bgroup \em et al.\egroup }{2020}]{mosavi2020comprehensive}
Amirhosein Mosavi, Yaser Faghan, Pedram Ghamisi, Puhong Duan, Sina~Faizollahzadeh Ardabili, Ely Salwana, and Shahab~S Band.
\newblock Comprehensive review of deep reinforcement learning methods and applications in economics.
\newblock {\em Mathematics}, 8(10):1640, 2020.

\bibitem[\protect\citeauthoryear{Oikonomou \bgroup \em et al.\egroup }{2023}]{oikonomou2023hybrid}
Katerina~Maria Oikonomou, Ioannis Kansizoglou, and Antonios Gasteratos.
\newblock A hybrid spiking neural network reinforcement learning agent for energy-efficient object manipulation.
\newblock {\em Machines}, 11(2):162, 2023.

\bibitem[\protect\citeauthoryear{Olkin and Pukelsheim}{1982}]{olkin1982distance}
Ingram Olkin and Friedrich Pukelsheim.
\newblock The distance between two random vectors with given dispersion matrices.
\newblock {\em Linear Algebra and its Applications}, 48:257--263, 1982.

\bibitem[\protect\citeauthoryear{Pacchiano \bgroup \em et al.\egroup }{2020}]{pacchiano2020learning}
Aldo Pacchiano, Jack Parker-Holder, Yunhao Tang, Krzysztof Choromanski, Anna Choromanska, and Michael Jordan.
\newblock Learning to score behaviors for guided policy optimization.
\newblock In {\em International Conference on Machine Learning}, pages 7445--7454, 2020.

\bibitem[\protect\citeauthoryear{Padakandla \bgroup \em et al.\egroup }{2020}]{padakandla2020reinforcement}
Sindhu Padakandla, KJ~Prabuchandran, and Shalabh Bhatnagar.
\newblock Reinforcement learning algorithm for non-stationary environments.
\newblock {\em Applied Intelligence}, 50(11):3590--3606, 2020.

\bibitem[\protect\citeauthoryear{Panaretos and Zemel}{2019}]{panaretos2019statistical}
Victor~M Panaretos and Yoav Zemel.
\newblock Statistical aspects of wasserstein distances.
\newblock {\em Annual review of statistics and its application}, 6:405--431, 2019.

\bibitem[\protect\citeauthoryear{Prabuchandran \bgroup \em et al.\egroup }{2021}]{prabuchandran2021change}
KJ~Prabuchandran, Nitin Singh, Pankaj Dayama, Ashutosh Agarwal, and Vinayaka Pandit.
\newblock Change point detection for compositional multivariate data.
\newblock {\em Applied Intelligence}, pages 1--26, 2021.

\bibitem[\protect\citeauthoryear{Rakelly \bgroup \em et al.\egroup }{2019}]{rakelly2019efficient}
Kate Rakelly, Aurick Zhou, Chelsea Finn, Sergey Levine, and Deirdre Quillen.
\newblock Efficient off-policy meta-reinforcement learning via probabilistic context variables.
\newblock In {\em International conference on machine learning(ICML)}, pages 5331--5340, 2019.

\bibitem[\protect\citeauthoryear{Sasaki and Yamashina}{2020}]{sasaki2020behavioral}
Fumihiro Sasaki and Ryota Yamashina.
\newblock Behavioral cloning from noisy demonstrations.
\newblock In {\em International Conference on Learning Representations}, 2020.

\bibitem[\protect\citeauthoryear{Sasaki}{2007}]{sasaki2007truth}
Yutaka Sasaki.
\newblock The truth of the f-measure.
\newblock {\em Teach tutor mater}, 1(5):1--5, 2007.

\bibitem[\protect\citeauthoryear{Schulman \bgroup \em et al.\egroup }{2015}]{schulman2015trust}
John Schulman, Sergey Levine, Pieter Abbeel, Michael Jordan, and Philipp Moritz.
\newblock Trust region policy optimization.
\newblock In {\em International conference on machine learning}, pages 1889--1897. PMLR, 2015.

\bibitem[\protect\citeauthoryear{Schulman \bgroup \em et al.\egroup }{2017}]{schulman2017proximal}
John Schulman, Filip Wolski, Prafulla Dhariwal, Alec Radford, and Oleg Klimov.
\newblock Proximal policy optimization algorithms.
\newblock {\em arXiv preprint arXiv:1707.06347}, 2017.

\bibitem[\protect\citeauthoryear{Schwarz \bgroup \em et al.\egroup }{2018}]{schwarz2018progress}
Jonathan Schwarz, Wojciech Czarnecki, Jelena Luketina, Agnieszka Grabska-Barwinska, Yee~Whye Teh, Razvan Pascanu, and Raia Hadsell.
\newblock Progress \& compress: A scalable framework for continual learning.
\newblock In {\em International Conference on Machine Learning(ICML)}, pages 4528--4537, 2018.

\bibitem[\protect\citeauthoryear{Shin \bgroup \em et al.\egroup }{2017}]{shin2017continual}
Hanul Shin, Jung~Kwon Lee, Jaehong Kim, and Jiwon Kim.
\newblock Continual learning with deep generative replay.
\newblock In {\em Advances in Neural Information Processing Systems(NIPS)}, pages 2990--2999, 2017.

\bibitem[\protect\citeauthoryear{Standley \bgroup \em et al.\egroup }{2020}]{standley2020tasks}
Trevor Standley, Amir Zamir, Dawn Chen, Leonidas Guibas, Jitendra Malik, and Silvio Savarese.
\newblock Which tasks should be learned together in multi-task learning?
\newblock In {\em International Conference on Machine Learning}, pages 9120--9132. PMLR, 2020.

\bibitem[\protect\citeauthoryear{Tiwari \bgroup \em et al.\egroup }{2023}]{tiwari2023consumer}
Prayag Tiwari, Abdullah Lakhan, Rutvij~H Jhaveri, and Tor-Morten Gronli.
\newblock Consumer-centric internet of medical things for cyborg applications based on federated reinforcement learning.
\newblock {\em IEEE Transactions on Consumer Electronics}, 2023.

\bibitem[\protect\citeauthoryear{Trabucco \bgroup \em et al.\egroup }{2022}]{trabucco2022anymorph}
Brandon Trabucco, Mariano Phielipp, and Glen Berseth.
\newblock Anymorph: Learning transferable polices by inferring agent morphology.
\newblock In {\em International Conference on Machine Learning}, pages 21677--21691. PMLR, 2022.

\bibitem[\protect\citeauthoryear{Van~Borkulo \bgroup \em et al.\egroup }{2022}]{van2022comparing}
Claudia~D Van~Borkulo, Riet van Bork, Lynn Boschloo, Jolanda~J Kossakowski, Pia Tio, Robert~A Schoevers, Denny Borsboom, and Lourens~J Waldorp.
\newblock Comparing network structures on three aspects: A permutation test.
\newblock {\em Psychological methods}, 2022.

\bibitem[\protect\citeauthoryear{Villani}{2009}]{villani2009optimal}
C{\'e}dric Villani.
\newblock {\em Optimal transport: old and new}, volume 338.
\newblock Springer, 2009.

\bibitem[\protect\citeauthoryear{Welch}{1990}]{welch1990construction}
William~J Welch.
\newblock Construction of permutation tests.
\newblock {\em Journal of the American Statistical Association}, 85(411):693--698, 1990.

\bibitem[\protect\citeauthoryear{Wydmuch \bgroup \em et al.\egroup }{2019}]{Wydmuch2019ViZdoom}
Marek Wydmuch, Micha{\l} Kempka, and Wojciech Ja\'skowski.
\newblock {ViZDoom} {C}ompetitions: {P}laying {D}oom from {P}ixels.
\newblock {\em IEEE Transactions on Games}, 11(3):248--259, 2019.
\newblock The 2022 IEEE Transactions on Games Outstanding Paper Award.

\bibitem[\protect\citeauthoryear{Xie \bgroup \em et al.\egroup }{2021}]{xie2021deep}
Annie Xie, James Harrison, and Chelsea Finn.
\newblock Deep reinforcement learning amidst continual structured non-stationarity.
\newblock In {\em International Conference on Machine Learning(ICML)}, pages 11393--11403, 2021.

\bibitem[\protect\citeauthoryear{Xu}{2019}]{xu2019approximation}
Lihu Xu.
\newblock Approximation of stable law in wasserstein-1 distance by stein’s method.
\newblock {\em The Annals of Applied Probability}, 29(1):458--504, 2019.

\bibitem[\protect\citeauthoryear{Yu \bgroup \em et al.\egroup }{2020}]{yu2020meta}
Tianhe Yu, Deirdre Quillen, Zhanpeng He, Ryan Julian, Karol Hausman, Chelsea Finn, and Sergey Levine.
\newblock Meta-world: A benchmark and evaluation for multi-task and meta reinforcement learning.
\newblock In {\em The Conference on Robot Learning (CoRL)}, pages 1094--1100, 2020.

\bibitem[\protect\citeauthoryear{Yu \bgroup \em et al.\egroup }{2021}]{yu2021conservative}
Tianhe Yu, Aviral Kumar, Yevgen Chebotar, Karol Hausman, Sergey Levine, and Chelsea Finn.
\newblock Conservative data sharing for multi-task offline reinforcement learning.
\newblock {\em Advances in Neural Information Processing Systems}, 34:11501--11516, 2021.

\bibitem[\protect\citeauthoryear{Yu \bgroup \em et al.\egroup }{2022}]{yu2022leverage}
Tianhe Yu, Aviral Kumar, Yevgen Chebotar, Karol Hausman, Chelsea Finn, and Sergey Levine.
\newblock How to leverage unlabeled data in offline reinforcement learning.
\newblock In {\em International Conference on Machine Learning}, pages 25611--25635. PMLR, 2022.

\bibitem[\protect\citeauthoryear{Zenke \bgroup \em et al.\egroup }{2017}]{zenke2017continual}
Friedemann Zenke, Ben Poole, and Surya Ganguli.
\newblock Continual learning through synaptic intelligence.
\newblock In {\em International Conference on Machine Learning (ICML)}, pages 3987--3995, 2017.

\bibitem[\protect\citeauthoryear{Zintgraf \bgroup \em et al.\egroup }{2019}]{zintgraf2019fast}
Luisa Zintgraf, Kyriacos Shiarli, Vitaly Kurin, Katja Hofmann, and Shimon Whiteson.
\newblock Fast context adaptation via meta-learning.
\newblock In {\em International Conference on Machine Learning}, pages 7693--7702. PMLR, 2019.

\end{thebibliography}
\end{document}